\pdfoutput=1

\documentclass[11pt]{article}

\usepackage{acl}

\usepackage{times}
\usepackage{latexsym}
\usepackage{booktabs}
\usepackage{multirow}
\usepackage{graphicx}
\usepackage{amsmath}
\usepackage{amssymb}
\usepackage{makecell}
\usepackage{hyperref}

\usepackage[T1]{fontenc}

\usepackage[utf8]{inputenc}

\usepackage{microtype}

\title{CleanGraph: Human-in-the-loop Knowledge Graph Refinement and Completion}

\author{Tyler Bikaun, Michael Stewart \and Wei Liu\\
  The University of Western Australia\\
  \texttt{tyler.bikaun@research.uwa.edu.au}\\}

\begin{document}
\maketitle

\begin{abstract}
This paper presents \textbf{CleanGraph}, an interactive web-based tool designed to facilitate the refinement and completion of knowledge graphs. Maintaining the reliability of knowledge graphs, which are grounded in high-quality and error-free facts, is crucial for real-world applications such as question-answering and information retrieval systems. These graphs are often automatically assembled from textual sources by extracting semantic triples via information extraction. However, assuring the quality of these extracted triples, especially when dealing with large or low-quality datasets, can pose a significant challenge and adversely affect the performance of downstream applications. CleanGraph allows users to perform Create, Read, Update, and Delete (CRUD) operations on their graphs, as well as apply models in the form of plugins for graph refinement and completion tasks. These functionalities enable users to enhance the integrity and reliability of their graph data. A demonstration of CleanGraph and its source code can be accessed at https://github.com/nlp-tlp/CleanGraph under the MIT License.
\end{abstract}

\section{Introduction}
\label{sec:introduction}

In the current data-centric era, where data is frequently referred to as the `new oil', knowledge graphs (KGs)\footnote{Alongside knowledge bases.}—structured representations of facts, encompassing semantically defined entities and relations—stand at the forefront of harnessing its value across various tasks, including question-answering \cite{huang2019knowledge}, information retrieval \cite{wise2020covid}, recommendation \cite{guo2020survey}, and reasoning \cite{chen2020review}.


Text-based knowledge graphs, a product of information extraction methods such as entity recognition, relation extraction, and entity linking \cite{Ji2022-zq}, have seen immense growth through automation, leading to expansive, encyclopedic graphs predominantly from high-quality web data \cite{bollacker2008freebase, Mitchell2018-gm} using techniques like distance supervision \cite{mintz2009distant}. Despite these advances, creating domain-specific knowledge graphs from diverse sources, including user-generated content or specialised fields like medicine, law enforcement, and engineering, has received less focus \cite{abu2021domain}. These domains present unique challenges due to resource scarcity, potential data quality issues, and the need for expert verification of factual accuracy. These difficulties have hindered the production of reliable, cost-effective knowledge graphs in these domains, leaving potential downstream applications underexplored compared to general-domain counterparts.

\begin{figure}[htbp]
    \centering
    \includegraphics[width=0.8\linewidth]{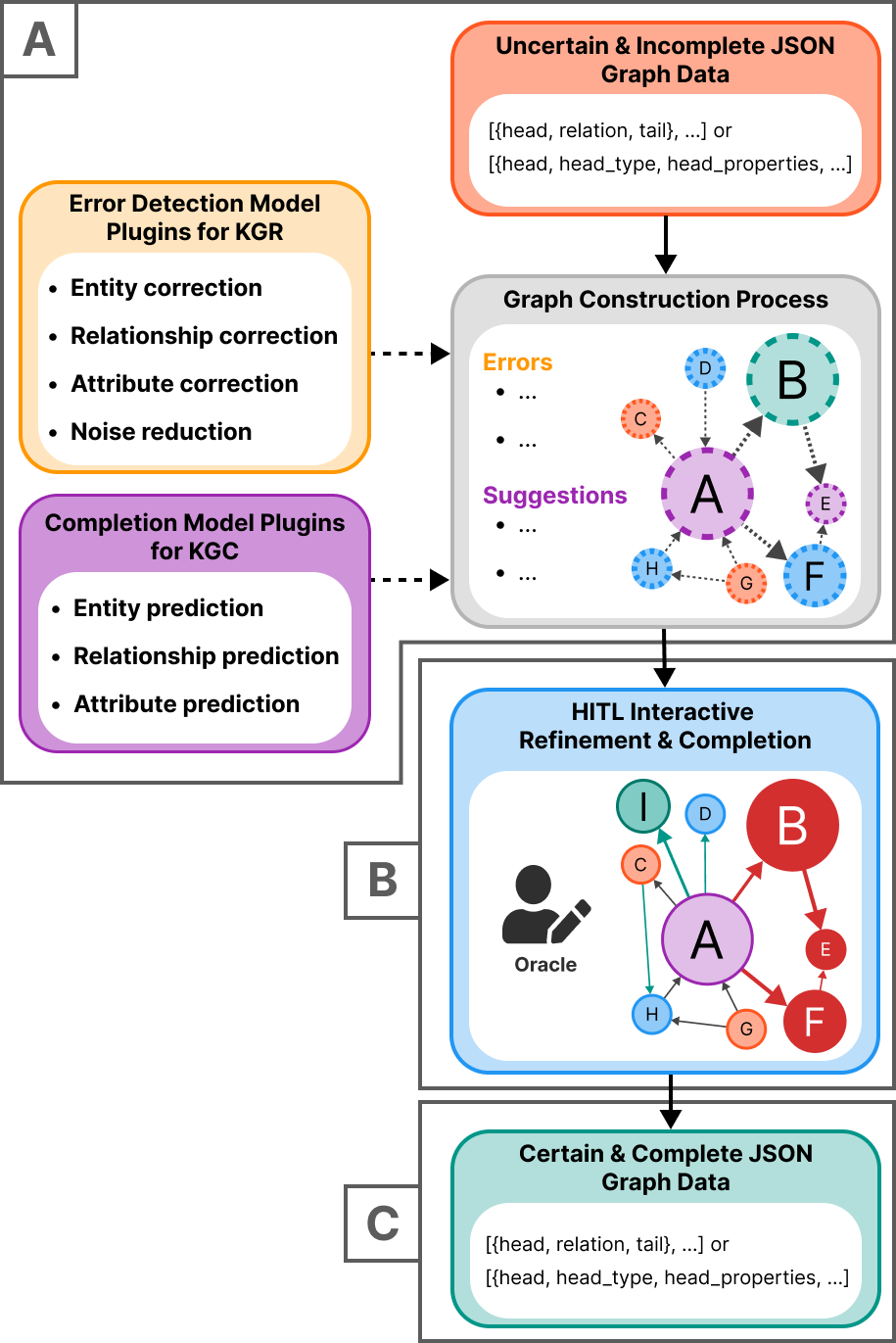}
    \caption{Schematic overview of the CleanGraph tool illustrating (A) graph data input, along with the use of optional model plugins for knowledge graph refinement (KGR) and completion (KGC), (B) the inclusion of human-in-the-loop (HITL) operations in the process, and (C) graph data output.}
    \label{fig:high_level_overview}
\end{figure}

While the need for high-quality and complete knowledge graphs predates the advent of large language models (LLMs) \cite{paulheim2017knowledge}, the widespread availability and capabilities of LLMs for zero or few-shot learning \cite{brown2020language} have further underscored this necessity. LLMs have democratised knowledge graph construction from various text corpora \cite{ye2022generative}, thus lowering the barriers to creating domain-specific graphs; however, they are not guaranteed to be factually correct. Consequently, maintaining the quality and completeness of these graphs remains paramount, especially for downstream applications like question-answering and recommendation, where factuality is crucial. As a result, there has been a concerted effort among researchers to improve knowledge graph quality and coverage via knowledge graph refinement (KGR) and completion (KGC) \cite{paulheim2017knowledge,rossi2021knowledge,Ji2022-zq}. 

However, there remains a deficiency in task-specific software capable of interactive, human-in-the-loop operations for these tasks. Presently, the available software predominantly supports visualisation and querying but lacks open accessibility or the capacity to incorporate KGR and KGC models easily. As a solution, such software could employ domain experts—or oracles—for quality assurance and validation, enhancing data accuracy before integration into downstream applications.

To address this gap, we introduce \textbf{CleanGraph}, a powerful yet intuitive web-based tool developed for introspection of, and interaction with, knowledge graphs\footnote{Represented as untyped and typed property graphs.} and results of KGR and KGC models (Figures \ref{fig:high_level_overview} and \ref{fig:ui_overview}). CleanGraph simplifies knowledge graph verification and refinement, making it user-friendly for technical and non-technical users. With built-in features that support visualisation and interaction, along with customisable KGR and KGC model plugins, CleanGraph assures an effortless quality control experience (Figure \ref{fig:high_level_overview}). This paper details the design, functionality, and architecture of CleanGraph, demonstrating its potential to enhance the integrity and reliability of knowledge graphs.


\section{Preliminaries - Knowledge graph refinement and completion}
\label{sec:preliminaries}
Knowledge graph refinement (KGR) and completion (KGC) are distinct, essential techniques used to enhance the quality and comprehensiveness of knowledge graphs \cite{paulheim2017knowledge, chen2020knowledge}. Despite extensive research in this domain, there remains a scarcity of accessible software allowing the direct involvement of an oracle in these processes. This section outlines both techniques, their subtasks, and their latest advancements.

\subsection{Knowledge graph refinement (KGR)}
KGR enhances the accuracy and relevance of the \textit{existing} facts within a knowledge graph \cite{paulheim2017knowledge}. This includes modifying or deleting incorrect or outdated entities (nodes), relationships (edges), or properties. The goal of KGR is to maintain the graph's accuracy, reliability, and overall quality. KGR tasks encompass:

\begin{itemize}
    \item \textbf{Entity correction:} Checking and rectifying incorrect entity representations (e.g., resolving ambiguities, synonyms, etc.) and types.
    \item \textbf{Relationship correction:} Examining and fixing erroneous relationships between entities, as well as types and directions.
    \item \textbf{Property correction:} Validating and correcting the property values associated with entities or relationships.
    \item \textbf{Noise reduction:} Removing irrelevant or redundant entities and relationships.
\end{itemize}

KGR challenges stem from the dynamic nature of knowledge leading to outdated information, misinterpretations, data input errors, and lack of context. Extensive research effort has been devoted to KGR, resulting in the development of statistical, machine learning, and deep learning techniques \cite{paulheim2017knowledge}. Current advancements in KGR focus on utilising knowledge representations like knowledge graph embeddings \cite{Ji2022-zq}.

\subsection{Knowledge graph completion (KGC)}
In contrast, KGC aims to \textit{fill in the missing gaps} within the knowledge graph \cite{chen2020knowledge}. This involves predicting and adding missing entities, relationships, or properties, enhancing the graph's comprehensiveness. KGC tasks include:

\begin{itemize}
    \item \textbf{Entity prediction:} Inferring and adding new missing entities from the graph.
    \item \textbf{Relationship prediction:} Discovering and adding missing relationships between existing entities.
    \item \textbf{Property prediction:} Predicting and adding missing entity or relationship properties.
\end{itemize}

KGC challenges arise from incomplete source data, the complexity of inferring unknown relations or properties, and the inherent uncertainty of predictions. Ensuring the accuracy of new information to prevent polluting the graph with incorrect data is also a major concern. Similar to KGR, KGC has received significant research attention, with recent advancements made in applying graph embeddings \cite{rossi2021knowledge,choudhary2021survey,Ji2022-zq}.

\begin{figure*}[ht]
  \centering 
  \includegraphics[width=\linewidth]{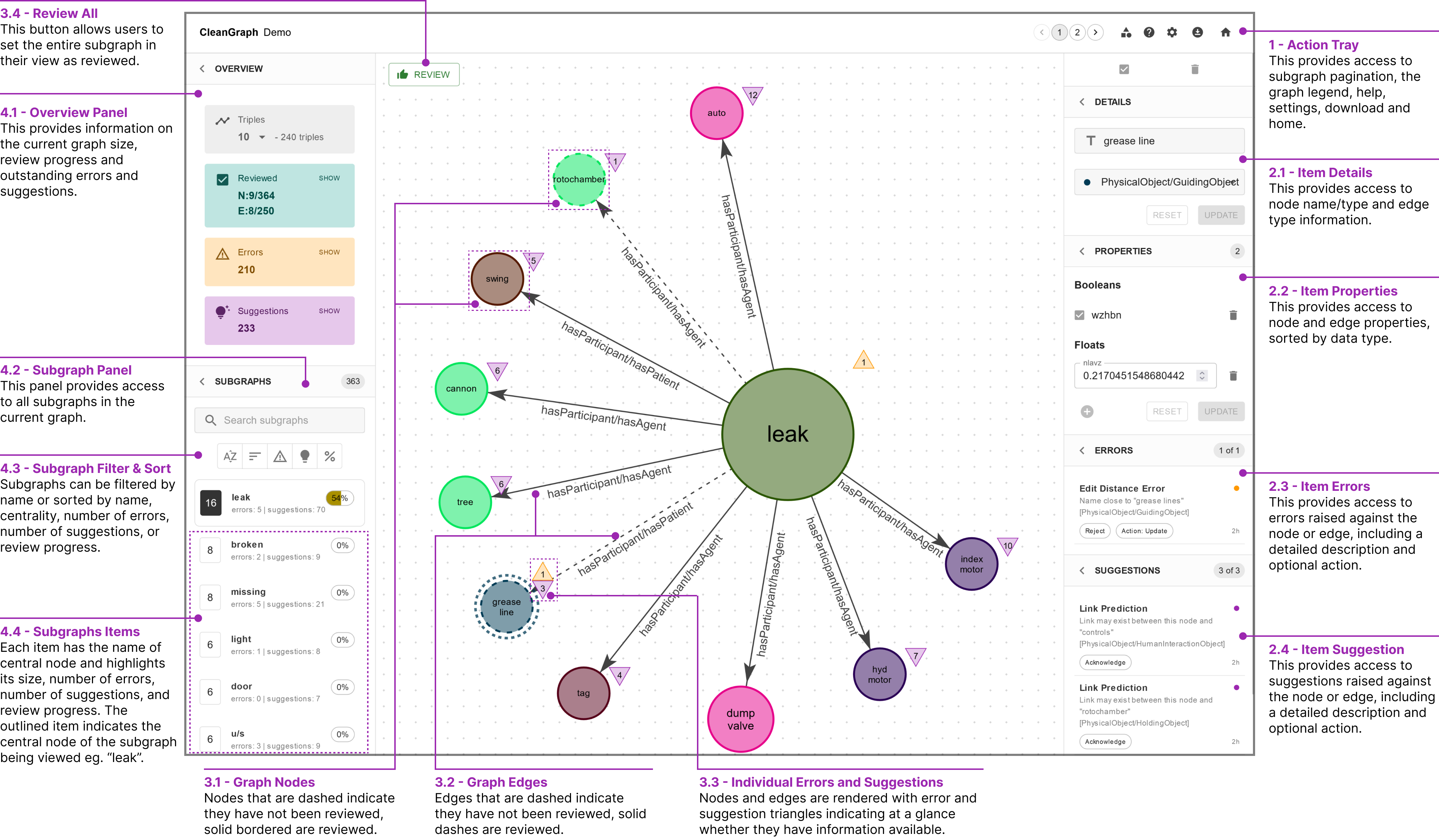}
  \caption{User interface of CleanGraph: Starting clockwise from the top right, (1) the action tray and subgraph pagination, (2) a secondary sidebar showing details, properties, errors, and suggestions for the chosen node or edge, (3) an interactive graph visualisation, and finally, (4) a primary sidebar displaying a progress overview and subgraphs.}
  \label{fig:ui_overview}
\end{figure*}




\section{Key features}
\label{sec:key_features}

The key features of CleanGraph, designed to simplify the introspection of knowledge graphs, are elaborated upon as follows.

\subsection{Graph creation and data structure}

Building a graph in CleanGraph is a user-friendly process. Users simply upload or input a JSON array of semantic triples, minimally in the \texttt{\{head, relation, tail\}} format\footnote{For an untyped graph.}. Models for error detection (KGR) and completion (KGC) can optionally be selected at this stage. To create more intricate property graphs, additional optional fields such as \texttt{types} and \texttt{properties} can be supplied (Figure \ref{fig:high_level_overview}A). Once the initial graph is constructed, any selected error detection and completion models are applied. This process populates the \texttt{errors} and \texttt{properties} fields of the relevant nodes and edges in the graph which are subsequently rendered in the user interface for human interaction (Figure \ref{fig:ui_overview}). 

\subsection{Graph visualisation and interaction}

\paragraph{Visualisation}

While viewing graph data in a tabular structure like \texttt{(head, relation, tail)} is sufficient for simple, untyped graphs, this approach is inefficient for larger, complex graphs. To address this, CleanGraph adopts a visually intuitive approach by representing graph data in its native format: as frequency-weighted nodes and edges (Figure \ref{fig:ui_overview}).

Graphs are displayed as directed acyclic graphs using a simulated force-directed layout, facilitated by the d3 visualisation library\footnote{Implemented via \textit{\href{https://github.com/vasturiano/react-force-graph}{react-force-graph}}.}. This layout, compared to tabular formats, allows users to visually parse the graph, identify patterns, and interact with nodes and edges easily through mouse actions and shortcuts. Users can quickly grasp the frequencies and types of nodes and edges, aided by colour coding. 

\begin{figure}[h!]
    \centering
    \includegraphics[width=0.5\linewidth]{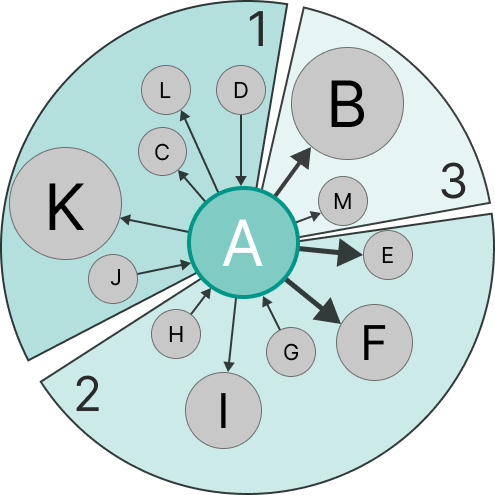}
    \caption{Illustration of CleanGraph's subgraph pagination process: A subgraph centred on the node (A) with 12 connected edges is split into 3 `pages' of 5 triples (size) for manageable viewing.}
    \label{fig:pagination_example}
\end{figure}

To manage information overload, CleanGraph partitions the entire graph into focused subgraphs; each centred on a single node. Users can then `paginate' through these subgraphs, effectively managing large graph data (Figure \ref{fig:pagination_example}). In addition, errors and suggestions identified by the EDM and CM plugins are displayed adjacent to the relevant nodes and edges, offering instant feedback to users (Figure \ref{fig:ui_overview}).

\paragraph{Interaction}

CleanGraph enables intuitive interaction with the graph. Users can manipulate the entire graph view through panning and zooming. Direct engagement with nodes and edges is possible via mouse clicks and shortcuts. Clicking on a node or edge reveals its details, properties, and any errors and suggestions (Figure \ref{fig:ui_overview}). Users can toggle the \texttt{reviewed} state of these elements or remove them from the graph entirely.


\subsection{Comprehensive graph CRUD operations}

CleanGraph is designed to provide comprehensive Create, Read, Update, and Delete (CRUD) operations across graphs, specifically accommodating the property graph format\footnote{This allows any arbitrary graph structure to be supported, even RDF-graphs, where the minimum amount of information is a head, relation and tail.}. It effectively handles nodes and edges, permitting them to support arbitrary key-value pair properties of any data type except objects and arrays. 

The core CRUD operations that are supported to make the manipulation of graph data\footnote{\textit{item(s)} refer to nodes and edges.} seamless include (1) the Addition of new nodes or edges to the graph, (2) Item review for error detection and graph refinement (Figure \ref{fig:ui_overview}), (3) Subgraph merging (section \ref{sssec:subgraph_merging}), (4) Acknowledgement and action on detected errors and suggested modifications (Figure \ref{fig:ui_overview}), (5) Reversal of edge direction to correct relational flow, and (6) Item deletion (section \ref{sssec:item_deletion}). These operations facilitate the integration of detected errors by KGR models and predictions made by KGC models into the graph. The subgraph merging and item deletion functionalities are in further detail below.

\subsubsection{Item deletion}
\label{sssec:item_deletion}
\begin{figure}[h]
    \centering
    \includegraphics[width=0.75\linewidth]{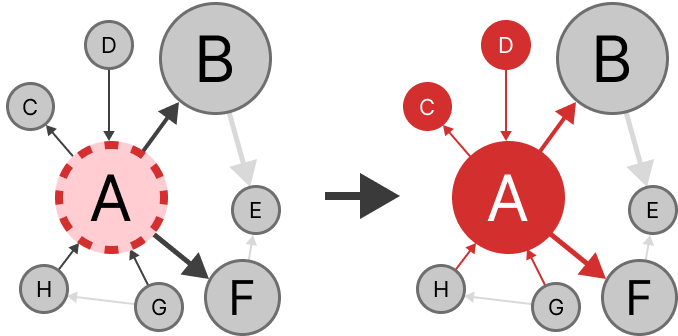}
    \caption{CleanGraph's 1-hop Item Deletion Illustrated: The removal of node (A) consequently eliminates all its corresponding edges and any nodes (C, D) that would become orphaned due to this operation.}
    \label{fig:1_hop_activation}
\end{figure}

CleanGraph facilitates KGR by enabling users to perform item deletions. To maintain the directed acyclic graph structure, it automatically propagates 1-hop edge removal and orphan node elimination when nodes or edges are deleted (Figure \ref{fig:1_hop_activation}).

\subsubsection{Subgraph merging}
\label{sssec:subgraph_merging}

\begin{figure}[h]
    \centering
    \includegraphics[width=0.75\linewidth]{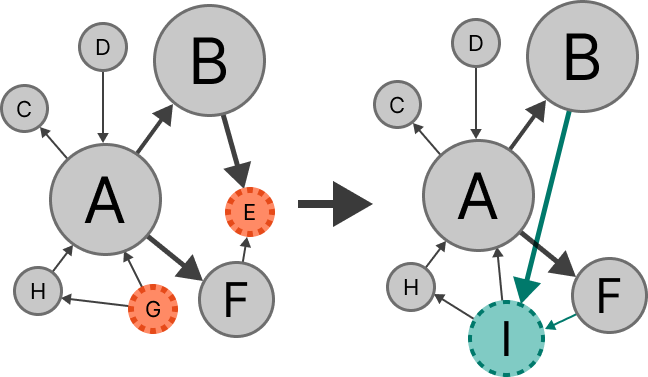}
    \caption{CleanGraph's Node Merge Illustrated: The merging of node (E) into (G) increments the node frequency and redistributes corresponding edges, resulting in a new node (I).}
    \label{fig:node_merge}
\end{figure}

CleanGraph's subgraph merging feature resolves entity errors. When a node's name or type changes, the system checks for potential conflicts with existing nodes. On detecting a conflict, it prompts the user to approve a proposed node merge. If approved, the respective subgraphs of the conflicting nodes merge, incorporating not only nodes and edges, but also associated properties, errors, suggestions, and frequencies for a comprehensive and accurate unification of the node data (Figure \ref{fig:node_merge}).

\subsection{Plugin architecture}
\label{ssec:plugin_architecture}

CleanGraph's plugin architecture enables flexible support for KGR and KGC tasks. By accommodating a broad spectrum of Error Detection Models (EDMs for KGR) and Completion Models (CMs for KGC), users can integrate any model that complies with the plugin interface, enabling seamless integration with the CleanGraph user interface (Figure \ref{fig:ui_overview}). Note: plugins are optional and only expedite the graph quality assurance process. 

\subsubsection{Authoring plugins}
CleanGraph's plugin authoring process is designed to promote collaboration and encourage the exchange of plugins via its open-source repository. These plugins may span heuristic, statistical, and deep-learning models, which we hope to evolve over time, leading to a shared resource community. To create a plugin, authors simply add a Python script that complies with the predefined plugin interface and adheres to the specified \href{https://docs.pydantic.dev}{Pydantic} input/output data models, into the server's \texttt{/plugins} directory. The input model is fed the graph as a set of semantic triples, and the output model returns a list of errors or suggestions, each containing relevant information and potential actions (defined and validated as Pydantic data models). These results are then rendered interactively on the user interface. Plugins are easily accessible from the user interface due to CleanGraph's plugin manager, ensuring reproducible usage across different graph construction processes.


\begin{figure}[htbp]
    \centering
    \includegraphics[width=\linewidth]{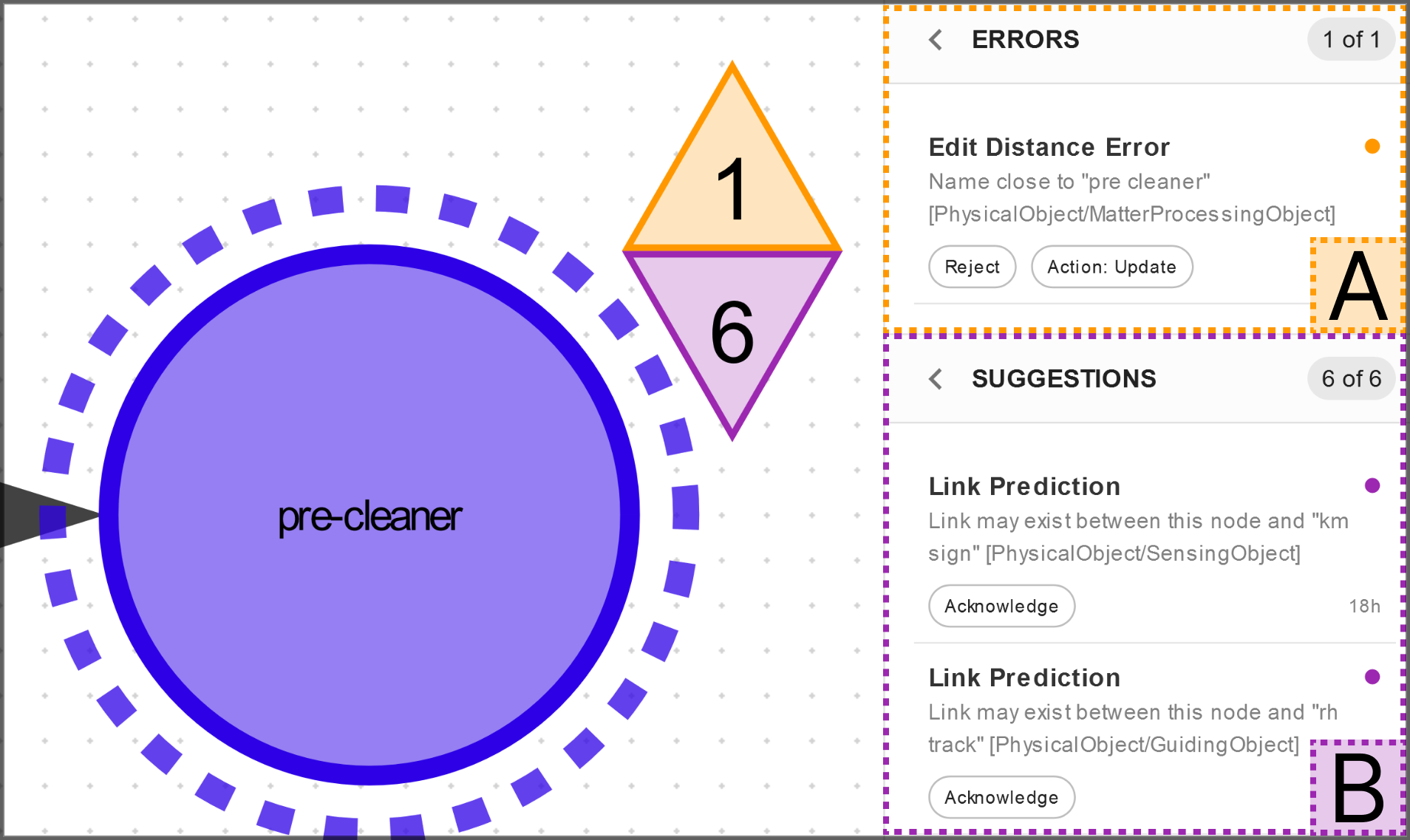}
    \caption{Display of CleanGraph's Error and Suggestion Features: (A) shows errors associated with a particular item (node), offering an optional corrective action (yellow triangle), while (B) presents informational suggestions (purple triangle). Both errors and suggestions can be acknowledged by the user.}
    \label{fig:ui_errors_and_suggestions_and_review_and_deactivated}
\end{figure}

\begin{table*}[t!]
    \footnotesize
    \caption{Comparative analysis of CleanGraph with existing tools for knowledge graph management and manipulation. Abbreviations: \textit{HITL}, \textit{KGR}, and \textit{KGC} refer to Human-in-the-Loop, Knowledge Graph Refinement, and Knowledge Graph Completion, respectively.}
    \begin{tabular}{p{4cm}cccccccc}
    \toprule
    Tool & Visualisation & Querying & \makecell{HITL\\CRUD} & \makecell{HITL\\KGR} & \makecell{HITL\\KGC} & \makecell{Open-\\Source} & \makecell{RDF-\\ based} & \makecell{Property \\Graph-based} \\
    \midrule
    \href{https://allegrograph.com/products/allegrograph/}{AllegroGraph} & \checkmark & \checkmark & - & - & - & - & \checkmark & - \\ 
    \href{https://blazegraph.com/}{Blazegraph} & - & \checkmark & - & - & - & - & \checkmark & - \\
    \href{https://neo4j.com}{Neo4J} & \checkmark & \checkmark & - & - & - & \checkmark & - & \checkmark \\
    \href{https://janusgraph.org/}{JanusGraph} & \checkmark & \checkmark & - & - & - & \checkmark &  - & \checkmark \\
    \href{https://www.tigergraph.com/}{TigerGraph} & \checkmark & \checkmark & - & - & - & - & - & \checkmark \\
    \midrule
    \textbf{CleanGraph} & \checkmark & - & \checkmark & \checkmark & \checkmark & \checkmark & - & \checkmark \\            
    \bottomrule
    \end{tabular}
    \label{tab:system_comparison}
\end{table*}

\subsubsection{Error detection models}
\label{sssec:error_detection_models}

Error detection model outputs, `errors', are displayed in the UI as easily identifiable `notifications' (Figures \ref{fig:ui_overview} and \ref{fig:ui_errors_and_suggestions_and_review_and_deactivated}A) attached to graph nodes and edges. Each notification can optionally include a corrective action, such as recommendations to merge nodes, update node types or labels, reverse edge directions, modify edge types, and update or delete node/edge properties. These actions assist the user in undertaking pertinent actions during the quality assurance process. CleanGraph defines a finite set of actions — \texttt{Update} and \texttt{Delete} — embodied in Pydantic data models, which can be freely used by EDM plugins. Nevertheless, actions are not mandatory; reviewers can simply `acknowledge' them within the UI.

\subsubsection{Completion models}
\label{sssec:completion_models}

Analogous to EDMs, completion model outputs, `suggestions', are also displayed in the UI similarly as `notifications' on nodes and edges (Figures \ref{fig:ui_overview} and \ref{fig:ui_errors_and_suggestions_and_review_and_deactivated}B). Suggestions notifications can have \texttt{Create} and \texttt{Update} actions, corresponding to the addition of new nodes/edges (\texttt{Create}) or updated node/edge properties (\texttt{Update}).

\section{System architecture}

CleanGraph is built using the FARM\footnote{FARM - FastAPI, React, MongoDB.} web application stack and comprises a server and a client. The server, implemented using \href{https://fastapi.tiangolo.com}{FastAPI} in Python, collaborates seamlessly with the client developed using a blend of HTML, Javascript (\href{https://react.dev}{React}), and CSS. The choice of a Python-based server is so that there is native support for KGR and KGC models typically implemented in Python libraries such as PyTorch and TensorFlow.

A NoSQL database, \href{https://www.mongodb.com}{MongoDB}, is used for data management, and stores graph data in four collections (\texttt{Graphs}, \texttt{Triples}, \texttt{Nodes}, and \texttt{Edges}). The choice of NoSQL over graph databases such as Neo4J owes to its support for graph-structured data and its ability to store arbitrarily structured data on nodes and edges. This flexibility circumvents the need for intricate representations often required by graph databases when dealing with complex data, such as nested arrays of complex objects for errors, suggestions, and properties.





\section{Comparison with existing tools}

Table \ref{tab:system_comparison} compares CleanGraph with existing tools for knowledge graph management. While most tools, such as AllegroGraph and Neo4J, are designed to handle large-scale knowledge graphs and lack Human-in-the-Loop (HITL) support, CleanGraph focuses on managing smaller text-based knowledge graphs with robust HITL functionalities.

Existing tools excel in visualising, storing, and querying extensive graph structures, but their support for HITL refinement and completion is often limited. Integration with external environments like Python is possible (e.g. \href{https://py2neo.org/2021.1/}{Py2Neo} for Neo4J), but these solutions generally fall short of a seamless iterative workflow involving human interaction. CleanGraph, however, offers visualisation, property graph-based support, and comprehensive HITL CRUD, KGR, and KGC functionalities. Its unique design emphasises flexibility and adaptability, enabling continuous human-involved refinement and completion.

It's important to recognize that this comparison may not be entirely fair, as the tools serve different purposes. Large graph databases aim to efficiently handle substantial data, while CleanGraph prioritises interactive, iterative development. This distinction, along with variations in open-source availability and feature sets, reflects the diverse goals and target audiences of these tools.

\section{Conclusion and future work}

This paper presented CleanGraph, a versatile and interactive tool designed for the refinement and completion of knowledge graphs. CleanGraph's unique contributions include providing comprehensive capabilities for graph CRUD operations and review that does not require any programming experience, alongside arbitrary knowledge graph refinement and completion models through its plugin infrastructure.

While CleanGraph is a robust, ready-to-use tool, it continues to evolve. Future enhancements include: i) expanding the range of available Error Detection and Completion plugins to incorporate graph embedding models, ii) supporting graph queries, iii) optimising performance for large-scale knowledge graphs, and iv) accommodating RDF-based semantic knowledge graphs.

\section*{Acknowledgements}

This research is supported by the Australian Research Council through the Centre for Transforming Maintenance through Data Science (grant number IC180100030), funded by the Australian Government. Additionally, Bikaun acknowledges funding from the Mineral Research Institute of Western Australia. Bikaun and Liu acknowledge the support from ARC Discovery Grant DP150102405.

\bibliography{custom}
\bibliographystyle{acl_natbib}

\end{document}